\documentclass{article}
\usepackage[a4paper, margin={3cm, 3cm}]{geometry}

\usepackage[utf8]{inputenc} %
\usepackage[T1]{fontenc}    %
\usepackage{hyperref}       %
\usepackage{url}            %
\usepackage{booktabs}       %
\usepackage{amsfonts}       %
\usepackage{nicefrac}       %
\usepackage{microtype}      %
\usepackage{xcolor}         %
\usepackage{amsmath}
\usepackage{listings}
\usepackage{subcaption}
\usepackage{problogconfig}
\usepackage{graphicx}
\usepackage{authblk}

\usepackage[inline]{enumitem}
\usepackage[normalem]{ulem}

\newcommand{\digit}[1]{\vcenter{\hbox{\includegraphics[height=10pt]{mnist/#1}}}}

\usepackage{amsthm}
\usepackage{mdframed}
\theoremstyle{definition}
\newmdtheoremenv[topline=false,rightline=false,bottomline=false]{example}{Example}

\usepackage{subcaption}
\newcommand\blfootnote[1]{%
  \begingroup
  \renewcommand\thefootnote{}\footnote{#1}%
  \addtocounter{footnote}{-1}%
  \endgroup
}

\makeatletter
\def\and{%
  \end{tabular}%
  \hskip 0.1em \@minus1cm%
  \begin{tabular}[t]{c}}%
\makeatother

\newcommand{\hwf}[1]{\vcenter{\hbox{\frame{\includegraphics[height=10pt]{hwf/#1.jpg}}}}}

\title{DeepStochLog: Neural Stochastic Logic Programming}

\author[1]{Thomas Winters\thanks{The authors contributed equally to this paper.}}
\author[1]{Giuseppe Marra$^*$}
\author[1]{Robin Manhaeve}
\author[1,2]{Luc De Raedt}
\affil[1]{KU Leuven, Dept. of Computer Science; Leuven.AI, B-3000 Leuven, Belgium, \texttt{\{firstname.lastname\}@kuleuven.be}}
\affil[2]{AASS, Örebro University, Sweden}

\date{}
\begin{document}

\maketitle
\blfootnote{Note that this paper will be extended in the future.}
\begin{abstract}
Recent advances in neural symbolic learning, such as DeepProbLog, extend probabilistic logic programs with neural predicates.
Like graphical models, these probabilistic logic programs define a probability distribution over possible worlds, for which inference is computationally hard. 
We propose DeepStochLog, an alternative neural symbolic framework based on stochastic definite clause grammars,  a type of stochastic logic program, which defines a probability distribution over possible derivations.
More specifically, we introduce neural grammar rules into stochastic definite clause grammars to create a framework that can be trained end-to-end.
We show that inference and learning in neural stochastic logic programming scale much better than for neural probabilistic logic programs.
Furthermore, the experimental evaluation shows that DeepStochLog achieves state-of-the-art results on challenging neural symbolic learning tasks.
\end{abstract}

\section{Introduction}\label{sec:intro}

The integration of neural and symbolic learning methods is high on the research agenda.
It is popular to use (variants of) logic programs to represent the available symbolic knowledge and use Prolog-like mechanisms to generate computation structures that can then be differentiated \cite{manhaeve2018deepproblog,dai,rocktaschel2017end,yang2020neurasp,tensorlog,ondrej,difflog}.
Several of these approaches also incorporate probability into these neural logic programming models, cf. \cite{deraedt2020starnesy}.
There are several ways to incorporate probability into logic programs \cite{de2015probabilistic}.
Most notably, one distinguishes probabilistic from stochastic logic programs (PLPs vs SLPs).
The more popular PLPs are based on a possible worlds semantics (the so-called distribution semantics), which extends probabilistic graphical models, while the SLPs are based on stochastic grammars. 
The difference can also be described as a random graph vs a random walk model.
So far, the emphasis in neurosymbolic computation has been on the PLP approach, especially \cite{yang2020neurasp,manhaeve2018deepproblog,ondrej,efthymia}, with only Tensorlog \cite{tensorlog} adopting the SLP semantics in an efficient but restricted Datalog or database setting that does not handle subsymbolic data such as images (see Section \ref{sec:rw} for details). 

To fill this gap, we introduce DeepStochLog, a neural stochastic logic programming approach.
It incorporates ideas from DeepProbLog such as the neural predicate.
The neural predicate encapsulates neural networks to cope, for instance, with subsymbolic data such as images. 
Without loss of generality, we base DeepStochLog on stochastic definite clause grammars (SDCGs) as this notation is not only easier to introduce and use, but also results in a sequence-based model.
SDCGs are a kind of probabilistic unification-based grammar formalism \cite{have2009stochasticdcg}.
However, SDCGs and SLPs are very closely related.
SDCGs can be directly translated and executed as SLPs, and all the concepts we introduce for SDCGs can directly apply to SLPs as well. 
More specifically, the key contributions of this paper are: 1) the introduction of the neural stochastic logic programming framework DeepStochLog;
2) the introduction of inference and learning algorithms (through gradient descent) for DeepStochLog programs; and
3) experimental results that show that DeepStochLog obtains state-of-the-art results on a number of 
challenging tasks for neural-symbolic computation and that it is also
several orders of magnitude faster than alternative approaches based on PLPs.

\section{Stochastic DCGs}
\label{sec:sdcg}

A context-free grammar (CFG) $G$ is a 4-tuple $(V,\Sigma,S, R)$, with $V$ the set of non-terminals, $\Sigma$ the set of terminals, $S \in V$ the starting symbol and $R$ a set of rewrite rules of the form $N \rightarrow S_1, ... , S_k$ where $N$ is a non-terminal, the $S_i$ are either terminals or non-terminals.
A probabilistic context-free grammar (PCFG) extends a CFG by adding probabilities to the rules $R$, i.e., the rules take the form $p_i :: N \rightarrow S_1, ... , S_k$, where $p_i$ is a probability.
Furthermore, the sum of the probabilities of rules with the same non-terminal $N$ on the left-hand side equals 1. 
We use list notation for sequences of terminals such as $[cat]$ and $[the, cat]$.
Whereas CFGs define whether a sequence can be parsed, PCFGs define a probability distribution over possible parses. This allows for the most likely parse to be identified.
An example PCFG, representing single digit additions, is shown on the left of Example \ref{ex:pcfgsdcg}.

Definite clause grammars (DCGs) are a well-known logic programming-based extension of CFGs \cite{pereira1980dcg}.
DCGs can represent context-sensitive languages and are unification-based. 
They differ from CFGs in that logical atoms are used instead of the non-terminals. 
An {\em atom} $a(t_1, ...,t_n)$ consists of a predicate $a$ of arity $n$ followed by $n$ terms $t_i$. 
Terms are either constants, logical variables or structured terms of the form $f(t_1, ... , t_k)$ with $f$ a functor and $t_j$ terms. 
The production rules are called definite clause grammar rules because they can be directly translated to a set of definite clauses (i.e., Horn clauses with exactly one positive literal) and can be executed as a Prolog program using SLD-resolution.
The right hand side of DCG rules are also allowed
to contain queries to Prolog predicates $q_i$ between curly brackets \texttt{\{$q_1(t_{1,1},...,t_{1,m_1}),...,q_n(t_{n,1},...,t_{n,m_n})$\}} to impose further constraints and perform additional computations during the inference process.
These are to be considered atoms as well.
DCGs use substitutions $\{V_1=t_1, ..., V_n=t_k\}$, which are sets of variable/term pairs, to unify atoms with heads from the rules.
Applying a substitution to an atom $a$ yields the atom $a\theta$ where all variables $V_i$ have been replaced by their corresponding terms $t_i$. $\theta$ is a unifier of an atom $s$ and an atom $u$ if and only if $s\theta = u\theta$.
For more information, see standard textbooks on logic programming such as \cite{Flach1994simplylogical,sterling1994artofprolog}.

Stochastic definite clause grammars (SDCGs) extend DCGs by associating probabilities to the rules, just like how PCFGs extend CFGs \cite{have2009stochasticdcg}.
As PCFGs, SDCGs require that the sum of the probabilities for the rules defining a single non-terminal predicate equals 1. SDCGs also correspond directly to stochastic logic programs (SLP) \cite{cussens2001parameterestimationslp,muggleton1996slp,muggleton2000learningslp}
which are well-known in the probabilistic (logic) programming community \cite{de2015probabilistic}. An example SDCG is shown in Example \ref{ex:pcfgsdcg} on the right.

\begin{example}%
\label{ex:pcfgsdcg}
A PCFG (left) and a similar SDCG(right) that constrains the result of the expression.\\
\begin{minipage}[t]{.43\linewidth}
    \begin{align*}
        0.5&::E \rightarrow N\\
        0.5&::E \rightarrow E,[+],N\\
        0.1&::N \rightarrow [0] \quad ...\quad0.1::N \rightarrow [9]
    \end{align*}
  \end{minipage}
  \begin{minipage}[t]{.55\linewidth}
\begin{align*}
0.5&::e(N) \rightarrow n(N)\\
0.5&::e(N) \rightarrow e(N1), [+], n(N2), \{N \;\text{is}\; N1 + N2\}\\
0.1&::n(0) \rightarrow [0]\quad ... \quad 0.1::n(9) \rightarrow [9]
\end{align*}
 \end{minipage}
\end{example}

The inference task in (S)DCGs consists of deriving a sequence of terminals from a goal (which often captures the starting symbol of the grammar). 
SLD-derivations are used for this. 
More formally, in an SLD-derivation for a DCG, a {\em goal} $g_1, ... ,g_n$ is a sequence where each $g_i$ is either a logical atom (a non-terminal) or a list containing terminals and logical variables. %
An SLD-derivation is shown in Example \ref{ex:derivation_example} and uses several resolution steps.
Applying {\em resolution} to a goal $g_1, ... ,g_n$ and a definite clause grammar rule $n \rightarrow t_1, ... , t_k$ yields the goal 
$g_1\theta, ... , g_{i-1} \theta, t_1 \theta, ... , t_k\theta, g_{i+1}\theta, ... , g_n\theta$ 
provided that $g_i$ is the leftmost atom in the goal (so $g_1, ... , g_{i-1}$ are terminal symbols), $\theta$ is the unifier of $g_i$ and $n$, i.e., $g_i \theta = n\theta$.\footnote{When a Prolog query $q$ is the first non-terminal to occur in a goal during the derivation process, the query is executed in Prolog possibly yielding an answer substitution $\theta$ such that $q\theta$ is true. For instance, in Example \ref{ex:pcfgsdcg}, there is the Prolog query $N~ \mathtt{is}~ N1 + N2$ which computes $N$ as the sum of $N1$ and $N2$.
In this paper, we assume that when such a query is called there is at most one answer substitution that is true. If there were more such substitutions, we would have to introduce a probability for such substitutions in the SDCG case, which unnecessarily complicates the semantics.}
We write this as $g_1, ... ,g_n \vdash t_1 \theta, ... , t_k\theta, s_2\theta, ... , s_n\theta$. 
A derivation $d(S)$ is then the repeated application $G \vdash G_1\theta_1 \vdash G_2\theta_1\theta_2 \vdash ... \vdash G_n\theta_1\theta_2 ... \theta_n$ of such resolution steps onto a goal $G$. 
We will write $G \vdash^* G_n$.
Successful derivations of a goal end in a sequence $T$ that consists only of terminal symbols, see Example \ref{ex:derivation_example} for an example. 
We will write that $d(S) = T$ and also say that $derives(S\theta,T)$ is true, with $\theta=\theta_1 \theta_2 ... \theta_n$ the answer substitution. 
A successful derivation corresponds to a proof.
The set of all possible proofs can be depicted using SLD-trees, see Figure~\ref{fig:inference_sld}.

The probability $P(d(G))$ of a derivation $d(G)$ is the product of the probabilities $\prod p_i^{m_i}$ of the rules $i$ used in the derivation with $m_i$ the number of times the rule $i$ was used.
An important difference between the probability of a parse in a PCFG and a derivation in an SDCG is that there can be a loss of probability mass in the latter whenever a derivation fails.
Derivations can fail when there are non-terminals in the goal that do not unify with the heads of any of the rules.
This is due to unification %
and is different from (P)CFGs, where non-terminals can always be resolved using rules for that non-terminal.
Non-terminating derivations can also lead to loss of probability mass.
Observe that every $G$ in this way induces a probability distribution $P_G$ over possible derivations $d(G)$. 
The goal can consist of one or more atoms, but in general for parsing, this will typically be the starting symbol or atom of the grammar.
This in turn lets us define the probability $P_G(derives(G,T))$
of a terminal sequence $T$  relative to the goal $G$ as $\sum_{d_i(G\theta)= T} P_G(d_i(G\theta))$, i.e. the sum of the probabilities of all derivations $G\theta$ for $G$ that result in the terminal sequence $T$ with answer substitution $\theta$. In a similar way, this allows to define the probability of an answer substitution $\theta$ relative to a goal $G$ and sequence $t$ as $P_G(derives(G\theta,t\theta))$ where $t$  could contain variables.   For ease of notation, when the goal is clear, we shall omit the subscript $G$.
Notice that if there are failing derivations, the total probability mass assigned to all sequences of terminals for a goal $G$ may be strictly less than 1. This is discussed at length by \cite{cussens2001parameterestimationslp}.
It is possible to obtain normalized probabilities by calculating the normalization constant, but this is computationally expensive.
We avoid this normalization in the present paper because in practice, the goal is often to find the most likely derivation $d_{max}(G,T) = \arg\max_{d(G) = T} P_G(d(G))$, non-normalized probabilities usually suffice. Notice that a SDCG defines a parametric probability distribution, where the parameters $p$ of the distribution are the vector of probabilities of the rules. When we need to refer to these probabilities we write  $P_G(derives(G,T); p)$.

\begin{example}[Derivations]
\label{ex:derivation_example}
Consider the following successful derivation using the SDCG in Example \ref{ex:pcfgsdcg} for the goal $G = [e(X)]$, the  answer substitution $\theta = \{X/2\}$ and the terminal sequence $T = [2, +, 0]$.\newline\noindent
\begin{minipage}{0.7\textwidth}
    \begin{align*}
        e(X) &\vdash e(N1), [+], n(N2), \{X \;\text{is}\; N1 + N2\} & \theta_1 = \{\} \\
        &\vdash n(N1), [+] , n(N2), \{X \;\text{is}\; N1 + N2\}  & \theta_2= \{\} \\
        &\vdash [2, +] , n(N2), \{X \;\text{is}\; 2 + N2\} &  \theta_3 = \{N1/2\} \\
        &\vdash [2,+,0], {2 \;\text{is}\; 2 + 0} & \theta_4 = \{X/2, N2/0\} 
    \end{align*}
    \end{minipage}
    \begin{minipage}{0.25\textwidth}
        \begin{align*}
         p = 0.5 \\
        \times 0.5 \\
        \times 0.1\\
        \times 0.1
    \end{align*}
    \end{minipage}
\end{example}

\section{DeepStochLog}

DeepStochLog integrates neural networks and SDCGs by introducing neural definite clause grammars (NDCG).
More formally, DeepStochLog allows for specifying an SDCG that additionally supports {\em neural definite clause grammar rules}, or neural rules for short. These are statements of the form:
\[
 nn( m,[I_1, ... , I_m],[O_1, ... ,O_L],[D_1, ... , D_L]) :: nt \rightarrow  g_1, ... , g_n
\]
where $nt$ is an atom, 
$g_1, ... , g_n$ is a goal,
and the $I_1 ... I_m$ and $O_1, ..., O_L$
are variables occurring in $g_1, ... , g_n $ and $nt$. The $D_i$ are unary predicates defining the domain of the output variables $O_i$.
The $nn$ declaration states that $m$ is a neural network that takes  the  variables $I_1, \ldots, I_m$ as input and outputs a probability distribution over output variables $O_1, ..., O_L$ (i.e. a probability distribution over the cross product of the domains specified by $D_i$). It thus maps an input substitution $\sigma$ for the variables $I_1, ..., I_m$ to a set of output substitutions $\theta_j$ with probability $p_j$.
The neural rule serves as a template. For every input substitution $\sigma$, the template  $(nt \rightarrow g_1, ... , g_n) \sigma$
defines the set of instantiated stochastic definite clause grammar rules 
$p_j:: (nt \rightarrow  g_1, ... , g_n)\sigma\theta_j$. 

\begin{example}[Neural definite clause grammar rules]
\label{ex:mnist_addition}
Consider the SDCG in Example~\ref{ex:pcfgsdcg}. We can substitute the $n(X)\rightarrow[X]$ rules with the following neural rule
\begin{equation*}
nn(mnist,[Mnist],[N],[digit]) :: n(N) \rightarrow [Mnist].
\end{equation*}
Here, the neural network called $mnist$ takes as input the $Mnist$ image and returns a probability for every possible number between 0 and 9, indicating how likely it is that every number is for the given MNIST image \cite{mnist}. The predicate $digit$ is defined as $digit(0), digit(1), ... , digit(9)$. 
Given the neural network and the input substitution $\sigma =\{ \texttt{Mnist =} \digit{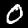} \}$ (which could be obtained through unification with the terminal sequence), the neural network could 
generate the output substitutions $\theta_0 = \{ N = 0\}$; ... ; $\theta_9 = \{ N = 9\}$; with probabilities $0.87$; ... ;$0.05$. 
Thus, the neural rule with the input substitution $\sigma =\{ \texttt{Mnist } = \digit{0} \}$ denotes the following set of grammar rules: 
$0.87::n(0) \rightarrow [\digit{0}];~~~\ldots;~~~$ $0.05::n(9) \rightarrow [\digit{0}]$
\end{example}
The neural rules are reminiscent of the neural predicates in DeepProbLog \cite{manhaeve2018deepproblog}, which also encapsulate a neural network that outputs a distribution over a number of alternatives. 
It is worth analyzing how a neural rule behaves w.r.t the neural inputs (e.g. images). In fact, a neural rule defines a probability distribution over the values of the output variables \textit{given} the neural inputs, whose distribution is not modeled in the program. This \textit{conditional} setting is akin to \textit{conditional} PCFGs \cite{riezler-etal-2002-parsing,sutton2006introduction} and it is a common modeling strategy in discriminative parsing\footnote{DeepStochLog could also be used to define generative grammars on subsymbolic inputs (e.g. images) if provided with neural models that can provide a joint probability distribution of both outputs and images. This is also discussed in \cite{manhaeve2021deepproblogjournal} but will not be further analyzed  in the current paper.}.

\section{Inference in DeepStochLog} \label{sec:inference}

\begin{figure}
    \centering
    \begin{subfigure}[t]{0.58\linewidth}
        \includegraphics[width=\linewidth]{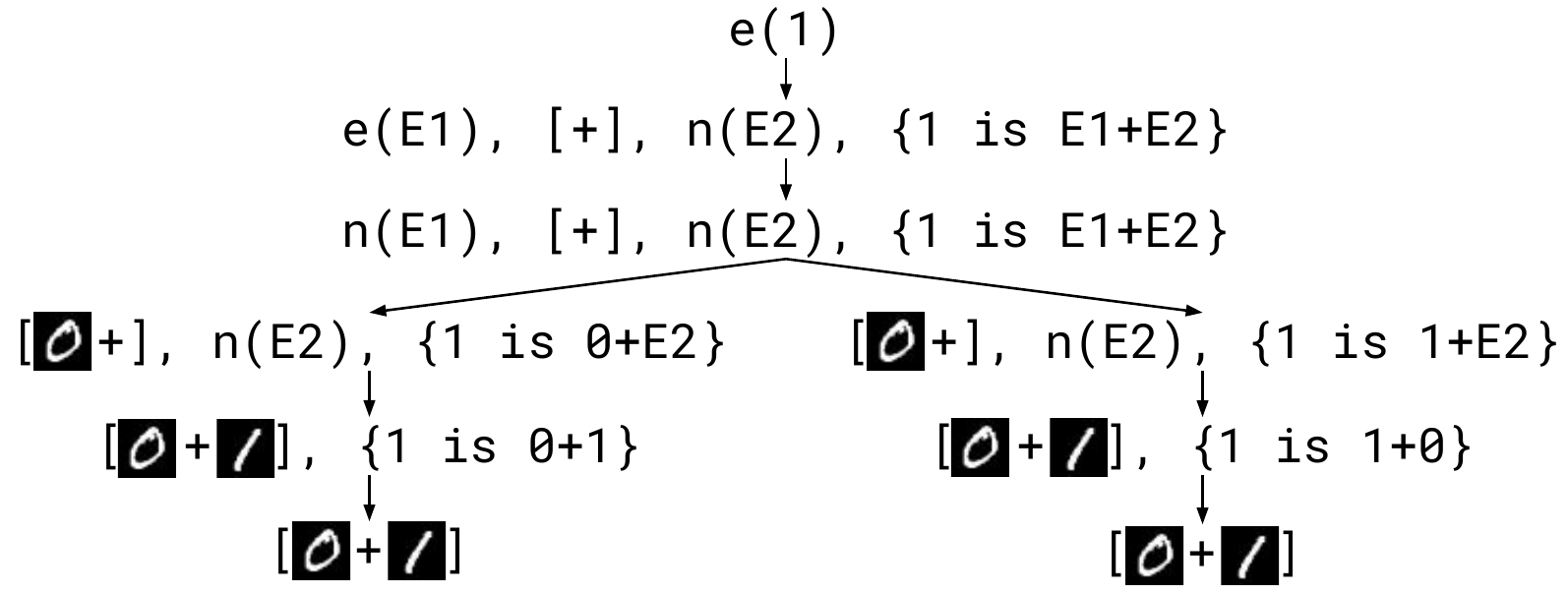}
        \caption{The SLD tree for $derives(e(1),[\digit{0}+\digit{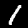}])$. Failing branches are omitted. 
        Notice that only the left-hand branch derives the correct parse of the images.
        }
        \label{fig:inference_sld}
    \end{subfigure}\hfill
    \begin{subfigure}[t]{0.39\linewidth}
        \includegraphics[width=\linewidth]{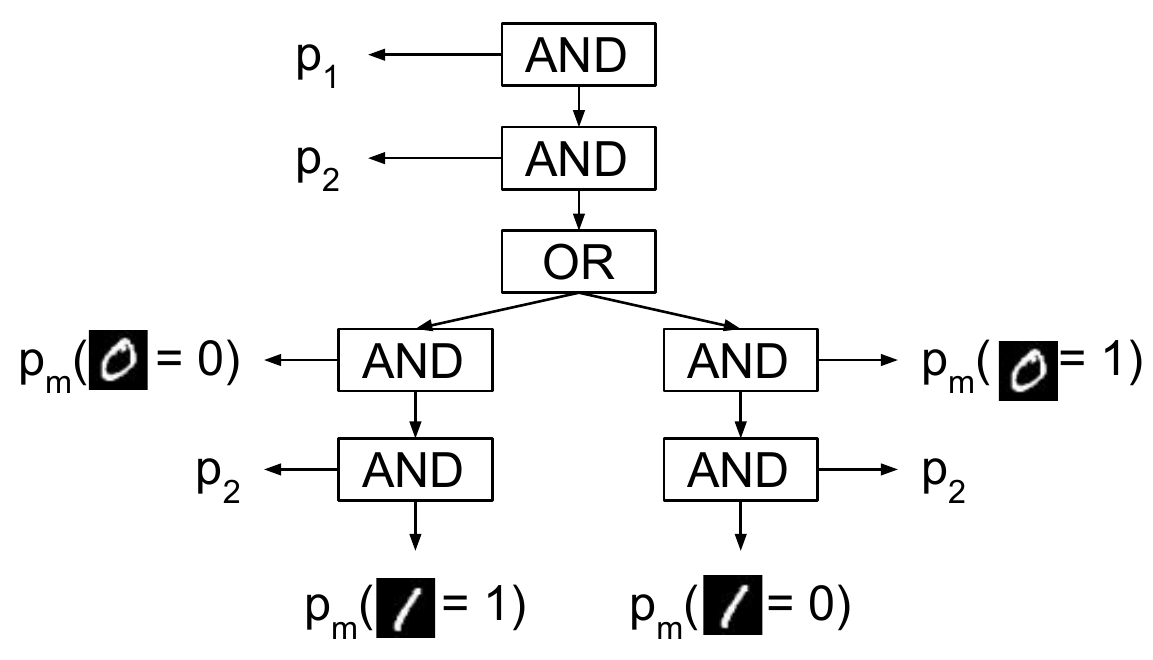}
        \caption{AND-OR circuit for $derives(e(1), [\digit{0}+\digit{1}])$ }
        \label{fig:inference_and_or}
    \end{subfigure}
    \caption{The different steps of inference on an example grammar.}
    \label{fig:inference_example}
\end{figure}

The goal of the inference is to compute the probability $P_G(derives(G,T))$ for a given goal $G$ and (possibly unknown) sequence $T$. This is divided into two steps, a logical and probabilistic one, which we now explain.
\paragraph{Logical inference}
Given a DeepStochLog program, a goal $G$ and a (potentially unknown) sequence $T$, logical inference uses resolution to answer $derives(G, T)$\footnote{This is sometimes called $phrase$ or $sentence$ in actual Prolog implementations and it requires an automatic syntactical translation of a DCG into Prolog. We show an example in Appendix \ref{app:translation-example}.}. This corresponds to finding all the possible derivations for $G$ that result in a terminal sequence $T$.
The resolution process is then turned into a compact AND-OR circuit, which represents all possible derivations and will be the input for the probabilistic inference.
The logical inference procedure is illustrated in Figure \ref{fig:inference_example}, where the SLD resolution tree for the given goal is on the left and its corresponding AND-OR circuit on the right. The translation to the AND-OR circuit is straightforward. It has exactly the same structure as the SLD-tree. For every resolution step with a rule $p_i:r_i$, an AND node is added. Furthermore,
for a normal SDCG rule, the corresponding probability $p_i$ is added,
and for a neural grammar rule, there is a call to the neural network that returns the probability $p_m$.
Whenever there are two (or more) branches in the SLD tree for a goal, an OR node is added. Notice that all the leaves are either probabilities given as parameters or the result of a neural call.

During SLD resolution, many identical intermediate goals may be proved multiple times which results in an explosion of inference time.
To avoid proving the same goals, we use SLG resultion~\cite{chen1996slgresolution}, which plays a similar role as the dynamic programming CYK algorithm for CFGs~\cite{kasami1966efficient}.
Tabling using SLG resolution is a standard logic programming technique that memoizes the answers of predicates by tabling the evaluations.
This technique is incorporated in Prolog implementations such as XSB, SWI-Prolog and Prism \cite{sato1997prism}.
The important difference with SLD resolution is that the results are not a single derivation tree, but rather a forest, where certain parts are re-used for multiple derivations thanks to tabled evaluations.
The effect of tabling is carried over to the creation of the AND-OR tree. Each time a derivation is reused from the table, its corresponding node is returned and linked to the new derivation. Thus, also the AND-OR circuit turns into a forest. 
When a goal admits a finite set of answers, we can resolve it only once and cache the corresponding AND-OR tree.

\paragraph{Probabilistic inference.}
With probabilistic inference, we refer to the task of calculating the probability $P(derives(G, T)) = \sum_{d(G\theta)= T} P(d(G\theta)) = \sum_{d(G\theta)= T} \prod_{r_i \in d(G\theta) } p_i^{m_i} $, 
i.e. the sum of the probabilities of all derivations for a given $G$ that result in a given terminal sequence $T$ and answer substitution $\theta$.
Thanks to SLG resolution and tabling, the shared sub-structure of many derivations is explicit in the AND-OR circuit obtained from the logical inference.
This dissipates the need for a specialized algorithm, like the \textit{inside} algorithm used in the probabilistic extension of CYK. 
Computing the probability $P(derives(G\theta, T))$ is just a bottom-up evaluation of the AND-OR circuit where AND-nodes are substituted by multiplications and OR-nodes by summations, i.e. compiling the logical circuit to an arithmetic circuit using the $(+, \times)$ semiring~\cite{kimmig2011aproblog}.
Analogously, the most probable derivation for the goal $G$ is found with the $(\max, \times)$ semiring.

\section{Learning in DeepStochLog}

Learning in DeepStochLog is achieved by optimizing the parameters of the neural networks and the parameters of the logic programs itself.
Let us consider a dataset of triples $\mathcal{D} = \{(G_i\theta_i, T_i, t_i)\}$, where $G_i$ is a goal, $\theta_i$ a substitution for $G_i$, $T_i$ a sequence of terminals and $t_i$ a target probability.
Let us also consider a DeepStochLog program parameterized by the vector $p$ of rule probabilities.
Learning in DeepStochLog is defined as the following optimization problem, with $\mathcal{L}$ being any differentiable loss function:
\begin{equation}
\label{eq:learning}
    \min_{p} \sum_{(G_i\theta_i, T_i, t_i) \in \mathcal{D}}  \mathcal{L}\bigg (P_G(derives(G_i\theta_i,T_i);p), t_i\bigg)
\end{equation}
Representing the dynamic programming computation for the \textit{inside} probability in terms of an arithmetic circuit has an important advantage. In fact, the corresponding computational graph is differentiable and the derivatives of the loss function $\mathcal{L}$ w.r.t. the probabilities $p$ can be carried out automatically using out-of-the-box differentiation frameworks. Moreover, when the probabilities $p$ are computed by a neural network 
as for a neural grammar rule, the gradients can be seamlessly backpropagated to the network to train its internal parameters.
We solve the learning problem using standard gradient descent techniques from deep learning, e.g. the Adam optimizer \cite{kingma2014adam}.

One interesting case is when the loss function $\mathcal{L}$ is the negative log-likelihood, as it brings DeepStochLog into the standard learning scenario for probabilistic grammars. Here, the optimization problem is usually carried out in the expectation-maximization (EM) framework. Given an \textit{inside} algorithm, a correspondent \textit{outside} algorithm is designed to extract the expected counts of the various grammar rules from data (E-step) and then the counts are used to update the probabilities (M-step). Most of the developed inside-outside algorithms are tailored to a specific formalism. However, the gradient descent approach of DeepStochLog on the negative log-likelihood is actually equivalent to the EM approach but it does not require the explicit definition of the corresponding outside algorithm. In fact, the gradients obtained by the backward pass through the AND-OR circuit have been shown to actually compute the outside probabilities (E-step), while the gradient descent step is used to update the parameters (M-step) \cite{salakhutdinov2003optimization,berg2010painless,eisner2016inside}.

\section{Evaluation}

\subsection{Research Questions}
The goal of our experiments is to answer the following questions:
\begin{itemize}
 \item[\textbf{Q1}] Does DeepStochLog reach state-of-the-art predictive performance on neural-symbolic tasks?

 \item[\textbf{Q2}] How does the inference time of DeepStochLog compare to other neural-symbolic frameworks and what is the role of tabling?
 
 \item[\textbf{Q3}] Can DeepStochLog handle larger-scale tasks?
 
 \item[\textbf{Q4}] Can DeepStochLog go beyond grammars and encode more general programs?

\end{itemize}

\subsection{Tasks}
\label{sec:tasks}

We specify the tasks used in this paper. Complete details are specified in Appendix~\ref{sec:details}.

\textbf{T1: MNIST Addition.}
In the MNIST Addition task \cite{manhaeve2018deepproblog}, the model is given two sequences of length $N$ of MNIST images containing handwritten images \cite{mnist}, each representing an N-digit number.
The task is to predict the sum of these numbers ($\digit{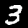}\digit{1}+\digit{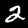}\digit{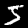}=56$). 
The training data only contains the two image sequences and the sum of the corresponding numbers, thus not providing the digit labels of the individual images.
The datasets for each digit length use all 60K images of MNIST images exactly once. 

\textbf{T2: Handwritten Formulas.}
In the Handwritten Formulas (HWF) task, the goal is to solve mathematical expressions, where both digits and operators (addition, subtraction, multiplication and division) are images of handwritten characters, like $\hwf{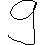}~\hwf{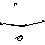}~\hwf{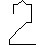}~\hwf{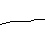}~\hwf{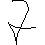}$.
Like T1, the data of T2 only contains the outcome of the expression and the sequence of images.
For this task, we use the Handwritten Formula (HWF) dataset, introduced in \cite{li2020ngs}. The dataset contains 10000 expressions of lengths 1, 3, 5 and 7. 
Unlike the original paper, we do not consider a curriculum learning setting here, and split the dataset into 4 separate parts by length.

\textbf{T3: Well-formed Parentheses.} We introduce the Well-formed Parentheses task, where the model is asked to recognize image sequences that represent well-formed parentheses.
Well-formed parentheses is a classic context-free grammar language where $\Sigma = \{(, )\}$, and $R=\{s\rightarrow() \rvert ( s ) \rvert s s \}$, i.e. all open brackets are closed in the right order.
As images, we use the zeros from MNIST as ``('' and ones as ``)'', and generate 1000 well-formed parenthesis sequences without labels as training data.
The goal is to predict the most probable parse of the bracket sequence.

\textbf{T4: Context-Sensitive Grammar.} Since DCGs support context-sensitive grammars, we created a dataset of 2000 image sequences representing the canonical context-sensitive grammar $a^n b^n c^n$.
Since each sequence of length $3n$ only has one valid parse, we increased the difficulty by allowing permutations such as $b^n a^n c^n$.
We also generated 2000 negative examples, i.e. random sequences of the form $a^k b^l c^m$, and permutations like $c^k a^l b^m$ such that 
$k$, $l$, $m$ are all larger than 1, sum to a multiple of 3 and are not all the same number. 
The goal of the task is to recognize whether the input sequence belongs to the first grammar or the second.

\textbf{T5: Semi-supervised classification in citation networks.} Given a set of scientific papers represented as bag-of-words and their citation network, the goal is to assign the correct class to a large test set of documents by having access only to the true labels of a small training set. The intuition is that one can infer the class of a paper not only by the features of the document but also by the class of its neighbors. This task is interesting from a neural symbolic perspective because one must be able to use both the features of the documents and the symbolic network. Two well-known datasets for this task are the Cora\footnote{On the Cora dataset, another common task is link prediction, which is a purely symbolic task (i.e. there are no subsymbolic features on the documents).} (2708 nodes and 5429 edges) and Citeseer (3327 nodes and 9228 edges) \cite{sen2008collective} .

\textbf{T6: Word Algebra Problems.}
In this task, a natural language text describes a word algebra problem (e.g, \textit{"Mark has 6 apples. He eats 2 and divides the remaining among his 2 friends. How many apples did each friend get?"}).
This dataset of this task contains 300 training instances and was introduced in \cite{roy2016solving}.
Each text contains 3 numbers, and all numbers have to be used exactly once in a formula containing addition subtraction, multiplication and division.
The task is to predict the right numerical answer to the expression implied by textual description.

\subsection{Results}
\label{sec:results}

For all experiments and metrics, we report the mean accuracy (or the mean most likely parse accuracy where applicable, i.e. \textbf{T1}, \textbf{T2} and \textbf{T3}) and standard deviation over 5 runs. We report ``timeout'' if a single of these 5 runs took more than 1 hour to execute.

\paragraph{Q1: Performance of DeepStochLog}
We first investigate whether DeepStochLog achieves  state-of-the-art results compared to similar neural-symbolic frameworks. %
Table~\ref{tab:mnist_accuracies} shows the result for the MNIST addition task (\textbf{T1}), for training and testing on lengths 1 to 4.
It shows that DeepStochLog performs similarly to the DeepProbLog \cite{manhaeve2018deepproblog} and NeurASP \cite{yang2020neurasp} frameworks but scales to larger sequences.
Table~\ref{tab:hwf_accuracies} shows the result on the HWF task (\textbf{T2}).
DeepStochLog performs similar to NGS and DeepProbLog for expressions of length 1 and 3.
Starting from expression length 5, it becomes infeasible to train DeepProbLog.
NGS \cite{li2020ngs} can still be trained, but for expression length 7, some runs fail to converge.
DeepStochLog, however, performs well for all expression lengths.
Table~\ref{tab:bracket_accuracies} shows the accuracy on task \textbf{T3}.
Here we can see that both DeepStochLog and DeepProbLog achieve high accuracy, but DeepStochLog reaches a slightly higher accuracy for a longer length.
For task \textbf{T6}, DeepStochLog and DeepProbLog achieve a similar accuracy of $94.8 \pm 1.1$ and $94.2 \pm 1.4$ respectively. We also compare to $\delta$4 \cite{riedel2017programming}, but the authors only report the maximum accuracy reached.
For all three frameworks, the maximum accuracy reached is $96.0$. To conclude, DeepStochLog is able to achieve similar or better performance compared to other state-of-the-art neural-symbolic frameworks.
 
\begin{table}[t]
\centering
\caption{The test accuracy (\%) on the MNIST addition (\textbf{T1}).}
\begin{tabular}{@{}lrrrr@{}}
\toprule
                      & \multicolumn{4}{c}{Number of digits per number (N)} \\
Methods               & 1                   & 2  &  3 & 4                  \\ \midrule
NeurASP  & $97.3 \pm 0.3$     & $93.9 \pm 0.7$  & timeout & timeout   \\
DeepProbLog  & $97.2 \pm 0.5$      & $95.2 \pm 1.7$     & timeout & timeout   \\
DeepStochLog  & $97.9 \pm 0.1$    & $96.4 \pm 0.1$   & $94.5 \pm 1.1$ & $92.7 \pm 0.6$ \\
\bottomrule
\end{tabular}

\label{tab:mnist_accuracies}
\end{table}

\begin{table}[t]
    \centering
    \caption{The accuracy (\%) on the HWF dataset (\textbf{T2}). }%
    \begin{tabular}{lrrrr}
    \toprule
    & \multicolumn{4}{c}{Expression length}\\
         Method & 1 & 3 & 5 & 7 \\
         \midrule
         NGS &  $90.2 \pm 1.6$ & $85.7 \pm 1.0$ & $91.7 \pm 1.3$ & $	20.4 \pm 37.2$\\
         DeepProbLog& $90.8 \pm 1.3$ & $85.6 \pm 1.1$ & timeout & timeout\\
         DeepStochLog &  $90.8 \pm 1.0$ & 
            $86.3 \pm 1.9$ &
            $92.1 \pm 1.4$ &
            $94.8 \pm 0.9$ 
             \\
         \bottomrule
    \end{tabular}
    
    \label{tab:hwf_accuracies}
\end{table}

\begin{table}[t]
    \centering
    \caption{The parse accuracy (\%) on the well-formed parentheses dataset (\textbf{T3}).}
    \begin{tabular}{lrrr}
    \toprule
     & \multicolumn{3}{c}{Maximum expression length}\\
         Method & 10 & 14 & 18  \\
         \midrule
         DeepProbLog  & $100.0 \pm 0.0$ & $99.4 \pm 0.5$ & $99.2 \pm 0.8$ \\
         DeepStochLog & $100.0 \pm 0.0$ & $100.0 \pm 0.0$ &  $100.0 \pm 0.0$ \\
         \bottomrule
    \end{tabular}
    
    \label{tab:bracket_accuracies}
\end{table} 

\begin{table}[t]
    \centering
    \caption{The accuracy (\%) on the $a^nb^nc^n$ dataset (\textbf{T4}).}
    \begin{tabular}{lrrr}
    \toprule
    & \multicolumn{3}{c}{Expression length}\\
         Method & 3-12 & 3-15 & 3-18  \\
         \midrule
         DeepProbLog  & $99.8 \pm 0.3$ 
 & timeout & timeout\\
         DeepStochLog & $99.4 \pm 0.5$ & $99.2 \pm 0.4$ & $98.8 \pm 0.2$     \\
         \bottomrule
    \end{tabular}
    \label{tab:anbncn_accuracies}
\end{table}

\textbf{Q2: DeepStochLog scalability}
We now investigate whether DeepStochLog is more scalable than similar neural-symbolic frameworks. 
First, we observe that in the tasks \textbf{T1}, \textbf{T2}, \textbf{T4} and \textbf{T5}, DeepStochLog scales to settings or datasets that are infeasible for the competitors. 
Next, in Table~\ref{tab:timing}, we compare the execution times for inference in task \textbf{T1}.
We selected 100 queries from the training data and we computed the average time required from the system to compute the probability of the query. We repeated the experiment for increasing number lengths.
DeepStochLog shows a huge gap over the competitors, especially for large numbers.
The advantage of DeepStochLog over the competitors is two-fold.
Firstly, DeepStochLog is natively implemented on top of SLG resolution and tabling, which plays a fundamental role in compactly representing derivations and SLD-trees.
We analyzed the impact of tabling in Table~\ref{tab:tabling}, where we show the comparison between SLD and SLG resolution in DeepStochLog. In particular, we compared the resolution time required to find all the possible answers for expressions of variable lengths (on task \textbf{T2}). 
Secondly, DeepStochLog is based on a random walk semantics which is computationally cheaper than the possible world semantics exploited by DeepProbLog and NeurASP. 

\begin{table}[t]
\begin{minipage}{0.45\linewidth}
\centering
\caption{\textbf{Q3} Accuracy (\%) of the classification on the test nodes on task \textbf{T5}}
\label{tab:results_citation}

    \begin{tabular}{l r r }
    \toprule
    \textbf{Method} & \textbf{Citeseer} & \textbf{Cora}\\
    \midrule
    ManiReg  & $60.1$ & $59.5$\\
    SemiEmb  & $59.6$ & $59.0$\\
    LP  & $45.3$ & $68.0$\\
    DeepWalk  & $43.2$ & $67.2$\\
    ICA  & $69.1$ & $75.1$ \\
    GCN  & $70.3$ & $81.5$ \\
    \midrule
    DeepProbLog & timeout & timeout \\
    DeepStochLog 
    & $65.0$
    & 
    $69.4$ \\
    \bottomrule
    \end{tabular}
\end{minipage}
\hfill
\begin{minipage}{0.50\linewidth}
\centering
\caption{\textbf{Q4} Parsing time in seconds (\textbf{T2}). Comparison of the DeepStochLog with and without tabling (SLD vs SLG resolution).}
\label{tab:tabling}
\begin{tabular}{@{}rrrr@{}}
\toprule
\textbf{Lengths} & \textbf{\# Answers} &\textbf{ No Tabling} & \textbf{Tabling}   \\ 
\midrule
1 & 10 & $0.067$ & $0.060$ \\
3 & 95 & $0.081$ & $0.096$\\
5 & 1066 & $3.78$ & $0.95$\\
7 & 10386 & $30.42$ & $10.95$\\
9 & 68298 & $1494.23$ & $132.26$\\
11 & 416517 &timeout& $1996.09$\\
    \bottomrule
\end{tabular}
\end{minipage}
\end{table}  

\begin{table}[t]
\centering
\caption{Inference times  in milliseconds for DeepStochLog, DeepProbLog and NeurASP on task \textbf{T1} for variable number lengths.}
\label{tab:timing}
\begin{tabular}{@{}lcccc@{}}
\toprule
Numbers Length      & 1
& 2 & 3 & 4 \\ \midrule
DeepStochLog &  $1.3 \pm  0.9$ & $2.3 \pm  0.4$ & $4.0 \pm  0.4$ & $5.7 \pm 1.8$ \\
DeepProbLog  &   $13.5 \pm 	3.0$  &  $36.0 \pm 0.5$  &  $199.7 \pm 14.0$        &   timeout                    \\
NeurASP  &  $9.2 \pm 1.4$ 
 &  $85.7 \pm 22.6$ 
 & $158.2 \pm  47.7$ & timeout \\
\bottomrule
\end{tabular}
\end{table}

\textbf{Q3: Larger scale relational datasets}
The complexity of many of the previous experiments comes from the large number of derivations for a single goal, while the number of subsymbolic inputs (e.g. images) in a single relational example was quite limited.
Here, we focus on task \textbf{T5}, i.e. semi-supervised classification in citation networks, where the complexity  mainly comes from the large number of elements of the unique relational example, i.e. the citation  network.
This task is usually out of the scope of (neural) PLP approaches due to the fact that there is a unique large relational example and the possible world semantics is prohibitive in this scenario.
We compare against the following baselines: label propagation (LP) \cite{zhu2003semi}, semi-supervised embedding (SemiEmb) \cite{weston2012deep}, manifold regularization (ManiReg) \cite{belkin2006manifold}, skip-gram based graph embeddings (DeepWalk) \cite{perozzi2014deepwalk}, ICA \cite{getoorICA} and GCN \cite{kipf2016semi}. All these baselines are specific to the semi-supervised classification task, while DeepStochLog is a much more general framework. We finally tried to compare with DeepProbLog, which, however, does not scale to the size of this problem due to the different probabilistic semantics. Results are reported in Table \ref{tab:results_citation}. DeepStochLog compares similarly or favorably to most of the other methods, even though it is the only one that has not been developed for the specific task. However, it still underperforms w.r.t. ICA and GCN. But these methods use extra knowledge as input to the classifier in the form of precomputed or learned features of the neighbors of a document, which is very useful for this task but not considered in the DeepStochLog experiment. %
Adding or learning relational features for input to the neural modules is, however, an interesting future direction.

\paragraph{Q4: General programs in DeepStochLog}
Even though DeepStochLog naturally represents grammars for parsing sequences, NDCGs with Prolog goals are a powerful formalism to express more complex relational problems and programs. Actually, both task \textbf{T5} and \textbf{T6} have been solved with programs that depart from the pure grammar formalism and are more like general logic programs. We provide the complete models in Appendix \ref{sec:details}. The main ingredients are (neural) empty production rules, sometimes referred to as non-consuming or $\epsilon$-production rules. They allow to take probabilistic decisions, including also neural networks, without consuming any element of the sequence, as shown in Example~\ref{ex:empty_productions}. This also shows that DeepStochLog has the full power of stochastic logic programs.

\begin{example}[Empty productions]
\label{ex:empty_productions}
We show a variant of the MNIST Addition problem using empty productions.
\begin{align*}
&nn(mnist,[X],[Y],[digit]) :: number(X,Y) \rightarrow []. \\
&addition(X,Y,N) \rightarrow number(X,N1), number(Y,N2), {N \;is\; N1 + N2}.
\end{align*}
This grammar will always produce the empty sequence but through the Prolog unification mechanism and the probabilistic modeling, we can express complex stochastic logic programs that include calls to neural networks.
\end{example}

\section{Related Works}
\label{sec:rw}
DeepStochLog is an expressive neural symbolic framework
whose distinguishing features are: 1) it is based on the expressive stochastic logic programming paradigm, which can express probabilistic programs (as in T5-T6) as well as probabilistic unification based grammars (T1-T4); 2) it can work with both symbolic and subsymbolic data such as images (as shown in T1-T4);
and 3) its inference and learning mechanism is based on SLG-resolution 
that naturally supports tabling, a form of dynamic programming (as shown in Q2). 

There are several strands of related research.
First, DeepStochLog is a neural logic programming language in the spirit of
DeepProbLog \cite{manhaeve2018deepproblog}, NeurASP \cite{yang2020neurasp}, the neural theorem prover \cite{rocktaschel2017end} and lifted relational neural networks (LRNNs) \cite{ondrej}.
The first two systems are based on a probabilistic possible world semantics, while DeepStochLog is based on stochastic grammars, which---as we have shown---scales much better (in part also due to the use of tabling).
The latter two approaches focus on Datalog (which cannot deal with function symbols) and use the logic to construct the neural network in a kind of knowledge based model construction approach.
Furthermore, they are neither probabilistic nor do they deal with subsymbolic inputs such as images.
Another related system is Tensorlog \cite{tensorlog}, which is based on stochastic logic programming.
While sharing their roots in SLPs, it is less expressive than DeepStochLog, as it considers only Datalog and predicates of arity 2.
While Tensorlog's implementation is fast thanks to being in terms of tensors, it
has only been applied to symbolic data. 

Second, DeepStochLog can be viewed as a neural-based grammar, similarly to Neural Grammars~\cite{dyer} and NGS~\cite{li2020ngs}. 
Neural Grammars have been introduced in the natural language community as an effective strategy to learn PCFGs. They are neural parameterizations of PFCG and it is possible to learn the structure of the grammar by enumerating a set of candidate rules and using neural networks to learn their probabilities. Differently from DeepStochLog, they are restricted to context-free grammars. 
Furthermore, Neural Grammars~\cite{dyer} do not consider subsymbolic inputs (as in all our tasks T1-T6). 
Different from the probabilistic interface of DeepStochLog, NGS uses backsearch, a greedy search that defines the backward feedback from the grammar to the neural nets. While this makes NGS very scalable, the backsearch must be defined \textit{per-program}, while DeepStochLog backpropagates evidence automatically through any NDCG. 
Neural Attribute Grammars~\cite{neuralattribute} integrate attribute grammars with neural networks.
While this is also an expressive grammatical framework, 
they are quite different from DeepStochLog in their approaches and applications, and also are not applied to subsymbolic data.

Third, many systems in the neural symbolic community \cite{LTN,SBR,ondrej} obtain differentiable logics by relaxing logical programs or theories using fuzzy logic and t-norms. While the shift in semantics from
probabilistic to fuzzy logic has known issues %
\cite{vankrieken}, fuzzy logic allows for more scalable systems as compared to probabilistic logic based on the possible world semantics. But by exploiting the stochastic grammars, DeepStochLog shows the same benefits as fuzzy logic in terms of computational complexity (i.e. no disjoint-sum problem required) by resorting to an alternative probabilistic semantics.

\section{Conclusions} \label{sec:conclusions}
We have introduced a novel and very expressive neural symbolic model based on stochastic logic programming, that allows to integrate symbolic knowledge with subsymbolic representations, that scales well, and gives state-of-the-art results on various neural symbolic computation tasks.

There are several limitations of DeepStochLog that we want to explore in further research.
First, DeepStochLog does not yet learn the structure of the rules, while the neural theorem prover \cite{rocktaschel2017end}, DiffLog \cite{difflog} and the neural grammars \cite{dyer} can all enumerate rules and then identify the most relevant ones. Second, DeepStochLog's inference could be further optimised by parallelization of the circuit using ideas from TensorLog \cite{tensorlog}.
Third, SLPs and hence, DeepStochLog, may lose probability mass due to failing derivations.
This can be addressed by normalizing and computing the partition function \cite{cussens2001parameterestimationslp}. It would be interesting to approximate the partition function and also to further speed up the inference by sampling or by searching for the k-best derivations. Finally, it would be interesting to explore the use of DeepStochLog as a generative model 
to generate sequences. 

\section{Acknowledgements}

We would like to thank Jessa Bekker for her helpful feedback and discussions throughout the whole project.
This work has received funding by the Research foundation - Flanders, the KU Leuven Research Fund, the European Research Council (ERC)
under the European Union’s Horizon 2020 research and innovation programme (grant agreement No [694980] SYNTH: Synthesising Inductive Data Models), the EU H2020 ICT48 project ``TAILOR'',
under contract \#952215; the Flemish Government under the ``Onderzoeksprogramma Artificiële Intelligentie (AI) Vlaanderen'' programme and the Wallenberg AI, Autonomous Systems and Software Program (WASP) funded by the Knut and Alice Wallenberg Foundation.
Thomas Winters is a fellow of the Research Foundation-Flanders (FWO-Vlaanderen, 11C7720N).
Robin Manhaeve is a SB PhD fellow of the Research Foundation-Flanders (FWO-Vlaanderen, 1S61718N).

\bibliographystyle{plain}
\bibliography{references.bib}

\begin{thebibliography}{10}

\bibitem{belkin2006manifold}
Mikhail Belkin, Partha Niyogi, and Vikas Sindhwani.
\newblock Manifold regularization: A geometric framework for learning from
  labeled and unlabeled examples.
\newblock {\em Journal of machine learning research}, 7(11), 2006.

\bibitem{berg2010painless}
Taylor Berg-Kirkpatrick, Alexandre Bouchard-C{\^o}t{\'e}, John DeNero, and Dan
  Klein.
\newblock Painless unsupervised learning with features.
\newblock In {\em Human Language Technologies: The 2010 Annual Conference of
  the North American Chapter of the Association for Computational Linguistics},
  pages 582--590, 2010.

\bibitem{riedel2017programming}
Matko Bošnjak, Tim Rockt{\"a}schel, and Sebastian Riedel.
\newblock Programming with a differentiable forth interpreter.
\newblock In {\em Proceedings of the 34th International Conference on Machine
  Learning}, volume~70, pages 547--556, 2017.

\bibitem{chen1996slgresolution}
Weidong Chen and David~S. Warren.
\newblock Tabled evaluation with delaying for general logic programs.
\newblock {\em J. ACM}, 43(1):20–74, January 1996.

\bibitem{tensorlog}
William~W Cohen, Fan Yang, and Kathryn~Rivard Mazaitis.
\newblock Tensorlog: Deep learning meets probabilistic databases.
\newblock {\em Journal of Artificial Intelligence Research}, 1:1--15, 2018.

\bibitem{cussens2001parameterestimationslp}
James Cussens.
\newblock Parameter estimation in stochastic logic programs.
\newblock {\em Machine Learning}, 44(3):245--271, 2001.

\bibitem{dai}
Wang-Zhou Dai, Qiuling Xu, Yang Yu, and Zhi-Hua Zhou.
\newblock Bridging machine learning and logical reasoning by abductive
  learning.
\newblock In H.~Wallach, H.~Larochelle, A.~Beygelzimer, F.~d\textquotesingle
  Alch\'{e}-Buc, E.~Fox, and R.~Garnett, editors, {\em Advances in Neural
  Information Processing Systems}, volume~32. Curran Associates, Inc., 2019.

\bibitem{deraedt2020starnesy}
Luc {De Raedt}, Sebastijan Dumančić, Robin Manhaeve, and Giuseppe Marra.
\newblock From statistical relational to neuro-symbolic artificial
  intelligence.
\newblock In Christian Bessiere, editor, {\em Proceedings of the Twenty-Ninth
  International Joint Conference on Artificial Intelligence, {IJCAI-20}}, pages
  4943--4950. International Joint Conferences on Artificial Intelligence
  Organization, 7 2020.
\newblock Survey track.

\bibitem{de2015probabilistic}
Luc De~Raedt and Angelika Kimmig.
\newblock Probabilistic (logic) programming concepts.
\newblock {\em Machine Learning}, 100(1):5--47, 2015.

\bibitem{SBR}
Michelangelo Diligenti, Marco Gori, and Claudio Sacca.
\newblock Semantic-based regularization for learning and inference.
\newblock {\em Artificial Intelligence}, 244:143--165, 2017.

\bibitem{LTN}
Ivan Donadello, Luciano Serafini, and Artur~S. d'Avila Garcez.
\newblock Logic tensor networks for semantic image interpretation.
\newblock In Carles Sierra, editor, {\em Proceedings of the Twenty-Sixth
  International Joint Conference on Artificial Intelligence, {IJCAI} 2017,
  Melbourne, Australia, August 19-25, 2017}, pages 1596--1602. ijcai.org, 2017.

\bibitem{dyer}
Chris Dyer, Adhiguna Kuncoro, Miguel Ballesteros, and Noah~A Smith.
\newblock Recurrent neural network grammars.
\newblock In {\em Proceedings of the 2016 Conference of the North American
  Chapter of the Association for Computational Linguistics: Human Language
  Technologies}, pages 199--209, 2016.

\bibitem{eisner2016inside}
Jason Eisner.
\newblock Inside-outside and forward-backward algorithms are just backprop
  (tutorial paper).
\newblock In {\em Proceedings of the Workshop on Structured Prediction for
  NLP}, pages 1--17, 2016.

\bibitem{Flach1994simplylogical}
Peter~A. Flach.
\newblock {\em Simply Logical Intelligent Reasoning by Example}.
\newblock John Wiley \& Sons, Inc., 1994.

\bibitem{have2009stochasticdcg}
Christian~Theil Have.
\newblock Stochastic definite clause grammars.
\newblock In {\em Proceedings of the International Conference RANLP-2009},
  pages 139--143, 2009.

\bibitem{kasami1966efficient}
Tadao Kasami.
\newblock An efficient recognition and syntax-analysis algorithm for
  context-free languages.
\newblock {\em Coordinated Science Laboratory Report no. R-257}, 1966.

\bibitem{kimmig2011aproblog}
Angelika Kimmig, Guy Van~den Broeck, and Luc De~Raedt.
\newblock An algebraic prolog for reasoning about possible worlds.
\newblock In {\em Proceedings of the AAAI Conference on Artificial
  Intelligence}, volume~25, 2011.

\bibitem{kingma2014adam}
Diederik~P Kingma and Jimmy Ba.
\newblock Adam: A method for stochastic optimization.
\newblock {\em arXiv preprint arXiv:1412.6980}, 2014.

\bibitem{kipf2016semi}
Thomas~N Kipf and Max Welling.
\newblock Semi-supervised classification with graph convolutional networks.
\newblock {\em arXiv preprint arXiv:1609.02907}, 2016.

\bibitem{mnist}
Yann Lecun, Léon Bottou, Yoshua Bengio, and Patrick Haffner.
\newblock Gradient-based learning applied to document recognition.
\newblock In {\em Proceedings of the IEEE}, pages 2278--2324, 1998.

\bibitem{li2020ngs}
Qing Li, Siyuan Huang, Yining Hong, Yixin Chen, Ying~Nian Wu, and Song-Chun
  Zhu.
\newblock Closed loop neural-symbolic learning via integrating neural
  perception, grammar parsing, and symbolic reasoning.
\newblock In {\em International Conference on Machine Learning}, pages
  5884--5894. PMLR, 2020.

\bibitem{getoorICA}
Qing Lu and Lise Getoor.
\newblock Link-based classification.
\newblock In Tom Fawcett and Nina Mishra, editors, {\em Machine Learning,
  Proceedings of the Twentieth International Conference {(ICML} 2003), August
  21-24, 2003, Washington, DC, {USA}}, pages 496--503. {AAAI} Press, 2003.

\bibitem{manhaeve2018deepproblog}
Robin Manhaeve, Sebastijan Dumancic, Angelika Kimmig, Thomas Demeester, and Luc
  De~Raedt.
\newblock Deepproblog: Neural probabilistic logic programming.
\newblock {\em Advances in Neural Information Processing Systems},
  31:3749--3759, 2018.

\bibitem{manhaeve2021deepproblogjournal}
Robin Manhaeve, Sebastijan Dumančić, Angelika Kimmig, Thomas Demeester, and
  Luc {De Raedt}.
\newblock Neural probabilistic logic programming in deepproblog.
\newblock {\em Artificial Intelligence}, 298:103504, 2021.

\bibitem{muggleton1996slp}
Stephen Muggleton.
\newblock Stochastic logic programs.
\newblock {\em Advances in inductive logic programming}, 32:254--264, 1996.

\bibitem{muggleton2000learningslp}
Stephen Muggleton.
\newblock Learning stochastic logic programs.
\newblock {\em Electron. Trans. Artif. Intell.}, 4(B):141--153, 2000.

\bibitem{neuralattribute}
Rohan Mukherjee, Dipak Chaudhari, Matthew Amodio, Thomas Reps, Swarat
  Chaudhuri, and Chris Jermaine.
\newblock Neural attribute grammars for semantics-guided program generation,
  2021.

\bibitem{pereira1980dcg}
Fernando~CN Pereira and David~HD Warren.
\newblock Definite clause grammars for language analysis—a survey of the
  formalism and a comparison with augmented transition networks.
\newblock {\em Artificial intelligence}, 13(3):231--278, 1980.

\bibitem{perozzi2014deepwalk}
Bryan Perozzi, Rami Al-Rfou, and Steven Skiena.
\newblock Deepwalk: Online learning of social representations.
\newblock In {\em Proceedings of the 20th ACM SIGKDD international conference
  on Knowledge discovery and data mining}, pages 701--710, 2014.

\bibitem{riezler-etal-2002-parsing}
Stefan Riezler, Tracy~H. King, Ronald~M. Kaplan, Richard Crouch, John~T.
  Maxwell~III, and Mark Johnson.
\newblock Parsing the {W}all {S}treet {J}ournal using a {L}exical-{F}unctional
  {G}rammar and discriminative estimation techniques.
\newblock In {\em Proceedings of the 40th Annual Meeting of the Association for
  Computational Linguistics}, pages 271--278, Philadelphia, Pennsylvania, USA,
  July 2002. Association for Computational Linguistics.

\bibitem{rocktaschel2017end}
Tim Rockt{\"a}schel and Sebastian Riedel.
\newblock End-to-end differentiable proving.
\newblock In {\em Advances in Neural Information Processing Systems},
  volume~30, pages 3788--3800, 2017.

\bibitem{roy2016solving}
Subhro Roy and Dan Roth.
\newblock Solving general arithmetic word problems.
\newblock In {\em Proceedings of the 2015 Conference on Empirical Methods in
  Natural Language Processing}, pages 1743--1752, 2015.

\bibitem{salakhutdinov2003optimization}
Ruslan Salakhutdinov, Sam~T Roweis, and Zoubin Ghahramani.
\newblock Optimization with em and expectation-conjugate-gradient.
\newblock In {\em Proceedings of the 20th International Conference on Machine
  Learning (ICML-03)}, pages 672--679, 2003.

\bibitem{sato1997prism}
Taisuke Sato and Yoshitaka Kameya.
\newblock Prism: a language for symbolic-statistical modeling.
\newblock In {\em IJCAI}, volume~97, pages 1330--1339, 1997.

\bibitem{sen2008collective}
Prithviraj Sen, Galileo Namata, Mustafa Bilgic, Lise Getoor, Brian Galligher,
  and Tina Eliassi-Rad.
\newblock Collective classification in network data.
\newblock {\em AI magazine}, 29(3):93--93, 2008.

\bibitem{difflog}
Xujie Si, Mukund Raghothaman, Kihong Heo, and Mayur Naik.
\newblock Synthesizing datalog programs using numerical relaxation.
\newblock In Sarit Kraus, editor, {\em Proceedings of the Twenty-Eighth
  International Joint Conference on Artificial Intelligence, {IJCAI} 2019,
  Macao, China, August 10-16, 2019}, pages 6117--6124. ijcai.org, 2019.

\bibitem{ondrej}
Gustav Sourek, Vojtech Aschenbrenner, Filip Zelezny, Steven Schockaert, and
  Ondrej Kuzelka.
\newblock Lifted relational neural networks: Efficient learning of latent
  relational structures.
\newblock {\em Journal of Artificial Intelligence Research}, 62:69--100, 2018.

\bibitem{sterling1994artofprolog}
Leon Sterling and Ehud~Y Shapiro.
\newblock {\em The art of Prolog: advanced programming techniques}.
\newblock MIT press, 1994.

\bibitem{sutton2006introduction}
Charles Sutton and Andrew McCallum.
\newblock An introduction to conditional random fields for relational learning.
\newblock {\em Introduction to statistical relational learning}, 2:93--128,
  2006.

\bibitem{efthymia}
Efthymia Tsamoura and Loizos Michael.
\newblock Neural-symbolic integration: {A} compositional perspective.
\newblock {\em CoRR}, abs/2010.11926, 2020.

\bibitem{vankrieken}
Emile van Krieken, Erman Acar, and Frank van Harmelen.
\newblock Analyzing differentiable fuzzy implications.
\newblock In Diego Calvanese, Esra Erdem, and Michael Thielscher, editors, {\em
  Proceedings of the 17th International Conference on Principles of Knowledge
  Representation and Reasoning, {KR} 2020, Rhodes, Greece, September 12-18,
  2020}, pages 893--903, 2020.

\bibitem{weston2012deep}
Jason Weston, Fr{\'e}d{\'e}ric Ratle, Hossein Mobahi, and Ronan Collobert.
\newblock Deep learning via semi-supervised embedding.
\newblock In {\em Neural networks: Tricks of the trade}, pages 639--655.
  Springer, 2012.

\bibitem{yang2020neurasp}
Zhun Yang, Adam Ishay, and Joohyung Lee.
\newblock Neurasp: Embracing neural networks into answer set programming.
\newblock In {\em Proceedings of the Twenty-Ninth International Joint
  Conference on Artificial Intelligence, IJCAI}, pages 1755--1762, 2020.

\bibitem{zhu2003semi}
Xiaojin Zhu, Zoubin Ghahramani, and John~D Lafferty.
\newblock Semi-supervised learning using gaussian fields and harmonic
  functions.
\newblock In {\em Proceedings of the 20th International conference on Machine
  learning (ICML-03)}, pages 912--919, 2003.

\end{thebibliography}

\newpage
\appendix

\section{Appendix}
\subsection{Data and Licenses}
\label{sec:licenses}
\begin{itemize}
    \item For tasks T1, T3 and T4, we used the MNIST dataset from \cite{mnist} to generate new datasets.
The MNIST dataset itself was released under the Creative Commons Attribution-Share Alike 3.0 license.
    \item The Handwritten Formula Recognition (HWF) dataset (used in T2) originates from \cite{li2020ngs}. %
    \item The Cora and Citeseer datasets (T5) are from \cite{sen2008collective}.
    \item The dataset of the Word Algebra Problem (T6) originates from \cite{roy2016solving}.
\end{itemize}

\subsection{Computational details} \label{sec:compute}

Inference time experiments are all executed on a MacBookPro 13 2020 (2.3 GHz Quad-Core Intel Core i7 and 16 GB 3733 MHz LPDDR4).

\section{Task Details}
\label{sec:details}
In this section we show the full DeepStochLog programs and additional experimental details for all tasks.

\subsection{MNIST Digit Addition}
\label{sec:addition-details}

For the MNIST addition problem, we trained the model for 25 epochs using the Adam optimizer with a learning rate of 0.001 and used 32 training terms in each batch for each digit length.
DeepStochLog program for single-digit numbers:

\begin{lstlisting}
digit(Y) :- member(Y,[0,1,2,3,4,5,6,7,8,9]).
nn(number, [X],[Y],[digit]):: number(Y) --> [X].
addition(N) -->  number(N1), number(N2), {N is N1+N2}.
\end{lstlisting}

DeepStochLog program for $L$-long digits numbers:
\begin{lstlisting}
digit(Y) :- member(Y,[0,1,2,3,4,5,6,7,8,9]).
nn(number, [X],[Y],[digit]):: number(Y) --> [X].
addition(N) -->  number(N1), number(N2), {N is N1+N2}.
0.5::multi_addition(N, 1) --> addition(N).
0.5::multi_addition(N, L) --> {L > 1, L2 is L - 1},
                                addition(N1), multi_addition(N2, L2), 
                                {N is N1*(10**L2) + N2}.
\end{lstlisting}

\subsection{Handwritten Mathematical Expressions}

For the Handwritten Mathematical Expression problem, we trained the model for 20 epochs using the Adam optimizer with a learning rate of 0.003 and a batch size of 2.
We used two separate, similar neural networks for recognising the numbers and the operators (see Table~\ref{tab:architectures}.
The encoder of the neural networks has a convolutional layer with 1 input channel, 6 output channels, kernel size 3, stride 1, padding 1, a ReLu, max pooling with kernel size 2, a convolutional layer with 6 inputs, 16 outputs, kernel size 3, stride 1, padding 1, a ReLu, a max pooling with kernel size 2, and a 2d dropout layer with $p=0.4$.
They then use two fully connected layers, one from 1936 to 128 with a ReLu, and one linear to the number of classes (10 and 4 respectively).
This problem is modeled as follows in DeepStochLog:

\begin{lstlisting}
digit(Y) :- member(Y,[0,1,2,3,4,5,6,7,8,9]).
operator_d(Y) :- member(Y,[plus, minus, times, div]).
term_switch_d(Y) :- member(Y,[0, 1, 2]).
e_switch_d(Y) :- member(Y,[0, 1, 2]).

nn(number, [X],[Y],[digit]):: number(Y) --> [X].
nn(operator, [X],[Y],[operator_d]) :: operator(Y) --> [X].
1 :: factor(N) --> number(N).

nn(term, [], [Y], [term_switch_d]) :: term(N) --> term_switch(N,Y).
0.33 :: term_switch(N, 0) --> factor(N).
0.33 :: term_switch(N, 1) --> term(N1), operator(times), factor(N2),
                            {N is N1 * N2}. 
0.33 :: term_switch(N, 2) --> term(N1), operator(div), factor(N2), 
                            {N2>0, N is N1 / N2}. 

nn(expression, [], [Y], [e_switch_d]) :: expression(N) --> e_switch(N,Y).
0.33 :: e_switch(N,0) --> term(N).
0.33 :: e_switch(N,1) --> expression(N1), operator(plus), term(N2), 
                        {N is N1 + N2}.
0.33 :: e_switch(N,2) --> expression(N1), operator(minus), term(N2), 
                        {N is N1 - N2}.
\end{lstlisting}

\subsection{Well-formed Parentheses}

We ran the well-formed parenthesis problem for 1 epoch using the Adam optimizer with a learning rate of 0.001 and a batch size of 4.
This problem is modelled as follows:

\begin{lstlisting}
bracket_d(Y) :- member(Y,["(",")"]).
s_switch_d(Y) :- member(Y,[0,1,2]).

nn(bracket_nn,[X], [Y], [bracket_d])::bracket(Y) --> [X].
nn(s_nn,[X],[Y],[s_switch_d])::s --> s_switch(Y).
0.33::s_switch(0) --> s, s.
0.33::s_switch(1) --> bracket("("), s, bracket(")").
0.33::s_switch(2) --> bracket("("), bracket(")").
\end{lstlisting}

\subsection{Context-Sensitive Grammar}

The model was trained on the canonical context-sensitive grammar $a^n b^n c^n$ problem for 1 epoch using the Adam optimizer with a learning rate of 0.001 and batch size 4.
This problem is modelled in DeepStochLog as follows:

\begin{lstlisting}
rep_d(Y):- domain(Y, [a,b,c]).
0.5:: s(0) --> akblcm(K,L,M),{K\=L; L\=M; M\=K}.
0.5:: s(1) --> akblcm(N,N,N).
akblcm(K,L,M) --> rep(K,A), rep(L,B), rep(M,C),{A\=B, B\=C, C\=A}.
0.5 :: rep(s(0), C) --> terminal(C).
0.5 :: rep(s(N), C) --> terminal(C), rep(N,C).
nn(mnist, [X], [C], [rep_d]) :: terminal(C) --> [X].
\end{lstlisting}

\subsection{Semi-supervised classification in citation networks}

The model was trained on the Cora and Citeseer datasets using the Adam optimizer with a learning rate of 0.01. We trained for 100 epochs and we selected the model corresponding to the epoch with the maximum accuracy on the validation set. %
We can easily and concisely express this problem in DeepStochLog as follows:

\begin{lstlisting}
class(Y) :- member(Y, [class0, class1, ..., class6]).
nn(classifier,[X],[Y],[class]):: doc_neural(X,Y) -->  [].

1/Na :: cite(a,b) --> [].
1/Na :: cite(a,c) --> [].
1/Nb :: cite(b,d) --> [].
...

0.5:: doc(X,Y) --> doc_neural(X,Y).
0.5:: doc(X,Y) --> cite(X, X1), doc(X1,Y).

s(X) --> doc(X,Y), [Y].
\end{lstlisting}

The program simply states that a document \texttt{X} is of class \texttt{Y} (and produce the terminal \texttt{Y}) either if a neural network classifies it so or if it is cited by one or more documents of the same class. Even if strange, many of these citation networks are actually cyclic. Therefore, we limited the depth of the derivations to a maximum value to avoid having a cyclical program. We also experimented with a variant of the program in which the probabilities of the rules expanding the \texttt{doc} predicate are trained separately for each class \texttt{Y}.

\subsection{Word Algebra Problem}

We trained the model on the Word Algebra Problem for 40 epochs using the Adam optimizer with a learning rate of 0.001 and a batch size of 32.
The DeepStochLog program is then modelled as follows:

\begin{lstlisting}
permute_d(Y) :- member(Y,[0,1,2,3,4,5]).
op(Y) :- member(Y, [plus,minus,times,div]).
swap_d(Y) :- member(Y, [no_swap, swap]).

nn(nn_permute, [X], [Y], [permute_d]):: nn_permute(X,Y) --> [].
nn(nn_op1, [X], [O], [op]):: nn_op1(X, O) --> [].
nn(nn_swap, [X], [S], [swap_d]):: nn_swap(X, S) --> [].
nn(nn_op2, [X], [O], [op]):: nn_op2(X, O) --> [].


0.1666::permute(0,A,B,C,A,B,C) --> [].
0.1666::permute(1,A,B,C,A,C,B) --> [].
0.1666::permute(2,A,B,C,B,A,C) --> [].
0.1666::permute(3,A,B,C,B,C,A) --> [].
0.1666::permute(4,A,B,C,C,A,B) --> [].
0.1666::permute(5,A,B,C,C,B,A) --> [].

0.5:: swap(no_swap,X,Y,X,Y) --> [].
0.5:: swap(swap,X,Y,Y,X) --> []. 

0.25:: operator(plus,X,Y,Z) --> [], {Z is X+Y}.
0.25:: operator(minus,X,Y,Z) --> [], {Z is X-Y}.
0.25:: operator(times,X,Y,Z) --> [], {Z is X*Y}.
0.25:: operator(div,X,Y,Z) --> [], {Y > 0, 0 =:= X mod Y, Z is X//Y}.

s(Out,X1,X2,X3) -->     [String],
                        nn_permute(String, Perm),
                        nn_op1(String, Op1),
                        nn_swap(String, Swap),
                        nn_op2(String, Op2),
                        permute(Perm,X1,X2,X3,N1,N2,N3),
                        operator(Op1,N1,N2,Res1),
                        swap(Swap,Res1,N3,X,Y),
                        operator(Op2,X,Y,Out).
\end{lstlisting}

\subsection{Neural network architectures} Table~\ref{tab:architectures} summarizes the neural network architectures used in the experiment. \textit{Conv(o,k)} is a convolutional layer with $o$ output channels and kernel size $k$. \textit{Lin(n)} is a fully-connected layer of size $n$. \textit{(Bi)GRU(h)} is a single-layer (bi-directional) GRU with a hidden size $h$.  \textit{MaxPool(k)} is a max-pooling layer with kernal size $k$, and \textit{Dropout(p)} a dropout layer with probability $p$.
A layer in bold means it is followed by a ReLU activation function. All neural networks end with a Softmax layer, unless otherwise specified.
\begin{table}[ht!]
\centering
\caption{Overview of the neural network architectures used in the experiments.}
\label{tab:architectures}
\begin{tabular}{@{}lll@{}}
\toprule
Task           & Network  & Architecture \\ \midrule
\textbf{T1} & number   &  MNISTConv,  \textbf{Lin(120)}, \textbf{Lin(84)}, Lin(10)\\
\textbf{T2} & number   & \textbf{Conv(6, 3)}, MaxPool(2), \textbf{Conv(16,3)}, MaxPool(2), Dropout(0.4),\\
&&\textbf{Lin(128)}, Lin(10)  \\
            & operator & \textbf{Conv(6, 3)}, MaxPool(2), \textbf{Conv(16,3)}, MaxPool(2), Dropout(0.4),\\
            &&\textbf{Lin(128)}, Lin(4)  \\
\textbf{T3} & bracket\_nn  &  MNISTConv,  \textbf{Lin(120)}, \textbf{Lin(84)}, Lin(2)\\
\textbf{T4} & mnist  &  MNISTConv,  \textbf{Lin(120)}, \textbf{Lin(84)}, Lin(3)\\
\textbf{T5} & classifier-Citeseer & \textbf{Lin(50)}, Lin(6) \\  
 & classifier-Cora & \textbf{Lin(50)}, Lin(7) \\
\textbf{T6}    & RNN      & Embedding(256), BiGRU(512), Dropout(0.5)*  \\
               & nn\_permute   & Lin(6)  \\
               & nn\_op1    & Lin(4)  \\
               & nn\_swap   & Lin(2)  \\
               & nn\_op2    & Lin(4) \\\bottomrule
\end{tabular}\\
{\small 
MNISTConv:  Conv(6,5), \textbf{MP(2,2)}, Conv(16,5), \textbf{MP(2,2)}*\\
AlexNetConv: \textbf{Conv(64, 11, 2,2)}, MP(3,2), \textbf{Conv(192, 5, 2)}, MP(3,2), \\\textbf{Conv(384, 3, 1)}, \textbf{Conv(256, 3, 1)}, \textbf{Conv(256, 3, 1)}, MP(3,2)*}\\
* Does not end with a Softmax layer.
\end{table}

\section{Translation Example of Neural Definite Clause Grammar to Prolog}
\label{app:translation-example}

Consider the following DeepStochLog program.
\begin{lstlisting}
digit(Y) :- member(Y, [0,1,2,3,4,5,6,7,8,9]).
op(Y) :- member(Y, [+,-]).

nn(mnist,[I],[N],[digit]) :: n(N) --> [I].
nn(operator,[I],[N],[op]) :: o(N) --> [I].

0.33::e(N) --> n(N).
0.33::e(S) --> e(E1), o(+), n(E2), {S is E1 + E2}.
0.33::e(S) --> e(E1), o(-), n(E2), {S is E1 - E2}.
\end{lstlisting}

It is translated into:
\begin{lstlisting}
digit(Y) :- member(Y, [0,1,2,3,4,5,6,7,8,9]).
op(Y) :- member(Y, [+,-]).

n(N, [I | X], [X]) :- nn(mnist,[I],[N]), digit(X).
n(N, [I | X], [X]) :- nn(mnist,[I],[O]), op(O).

e(N, A, B) :- n(N, A, B), p(0.33).
e(S, A,D) :- e(E1, A, B), o(+, B, C), n(E2, C, D), S is E1 + E2, p(0.33).
e(S, A,D) :- e(E1, A, B), o(-, B, C), n(E2, C, D), S is E1 - E2, p(0.33).
\end{lstlisting}

Both the calls to \texttt{nn} and \texttt{p} are considered always \texttt{true} during logical inference. During evaluation of the arithmetic circuit, \texttt{nn(mnist,[I],[O])} returns the probability of the output \texttt{O} when provided with the image \texttt{I} as input, while \texttt{p(x)} constantly returns the probability \texttt{x}. It is easy to see that any derivation of the Prolog program always end with either a \texttt{nn} or a \texttt{p} call, which constitute the leaves of the correspondent arithmetic circuit.

\end{document}